\pdfoutput=1
\documentclass[letterpaper, 10 pt, conference]{ieeeconf}  % Comment this line out if you need a4paper

\IEEEoverridecommandlockouts                              % This command is only needed if 
\usepackage{cite}
\usepackage{colortbl}
\usepackage{xcolor}
\usepackage{graphicx}
\usepackage{wrapfig}
\usepackage[font=footnotesize]{subcaption}
\usepackage{changepage}
\usepackage{amsmath}
\usepackage{algorithm}
\usepackage{amssymb}
\usepackage{amsmath}
\usepackage{algorithm}
\usepackage[noend]{algpseudocode}
\usepackage{resizegather}

\usepackage[utf8]{inputenc}
\usepackage[english]{babel}

\makeatletter
\def\BState{\State\hskip-\ALG@thistlm}
\makeatother

\newlength\myindent
\title{\LARGE \bf
Personalizing User Engagement Dynamics \\ in a Non-Verbal Communication Game for Cerebral Palsy
}

\author{Nathaniel Dennler$^{1}$, Catherine Yunis$^{2}$, Jonathan Realmuto$^{3}$, \\ Terence Sanger$^{3}$, Stefanos Nikolaidis$^{1}$,  and Maja Matari\'c$^{1}$% <-this % stops a space
\thanks{$^{1}$Nathaniel Dennler, Stefanos Nikolaidis, and Maja Matari\'c are from the Computer Science Department at the University of Southern California in Los Angeles, CA
        {\tt\small \{dennler, nikolaid, mataric\} @usc.edu}}%
        
\thanks{$^{2}$Catherine Yunis is from the Biomedical Engineering Department at the University of Southern California in Los Angeles, CA
        {\tt\small cyunis@usc.edu}}%
\thanks{$^{3}$Jonathan Realmuto and Terence Sanger and are from the Electrical Engineering and Computer Science Department at University of California, Irvine in Irvine, CA
        {\tt\small \{jrealmut, tsanger\}@uci.edu}}%
}

\begin{document}

\maketitle
\thispagestyle{empty}
\pagestyle{empty}

\begin{abstract}
    Children and adults with cerebral palsy (CP) can have involuntary upper limb movements as a consequence of the symptoms that characterize their motor disability, leading to difficulties in communicating with caretakers and peers. We describe how a socially assistive robot may help individuals with CP to practice non-verbal communicative gestures using an active orthosis in a one-on-one number-guessing game. We performed a user study and data collection with participants with CP; we found that participants preferred an embodied robot over a screen-based agent, and we used the participant data to train personalized models of participant engagement dynamics that can be used to select personalized robot actions. Our work highlights the benefit of personalized models in the engagement of users with CP with a socially assistive robot and offers design insights for future work in this area.
    
\end{abstract}

\section{Introduction}

% Socially Assistive Robots (SAR) is a promising subfield of Human-Robot Interaction that has a large potential to augment care-giving through social interaction ~\cite{feil2005defining,mataric2016socially}.

Cerebral palsy (CP) is one of the most prevalent motor disorders in children \cite{oskoui2013update}, affecting around 0.2\%-0.3\% of all live births in the United States \cite{winter2002trends}. The main symptom of CP is involuntary muscle contractions that lead to repetitive movements \cite{sanger2004toward} which can greatly affect a  child's ability to communicate with caregivers and peers \cite{hidecker2011developing}. This symptom necessitates the use of active orthoses to facilitate proactive communication and to aid in motor rehabilitation \cite{realmuto2019robotic}.  

Retraining motor skills, however, requires repetitive and lengthy sessions to be effective \cite{buitrago2020motor}. In children especially, this can lead to disengagement with the therapeutic activity, negatively affecting functional outcomes. Thus, we aim to facilitate engaging therapeutic activities through the development of an engaging game between a participant and a socially assistive robot (SAR)~\cite{feil2005defining}\cite{mataric2016socially}. The robot encourages the participant to perform (and therefore practice) non-verbal communicative gestures while providing social reinforcement as the participant makes progress in the game.

This work explores the effect of both physical and social factors of the interaction design on the ability to effectively engage participants. Physically, we investigate the embodiment of the agent that delivers the game. Several recent works have described the positive effect of strongly embodied agents, such as robots, on the engagement of participants over weakly embodied agents, such as computers (see review by Deng et al. \cite{deng2019embodiment}). Socially, we investigate how the feedback provided by the agent throughout the game affects participant engagement. Understanding how robots can effect engagement dynamics is an under-explored area of human-robot interaction (HRI) \cite{oertel2020engagement}.

Both physical and social factors are investigated through a user study of participants with CP.  We found that participants preferred interacting with the SAR compared to a screen-based agent but did not observe any significant differences in engagement levels between the two conditions, which we attribute to individual differences in how participants responded to the robot's actions, as detailed in Section~\ref{clusters}. To explore user engagement further, we then developed a probabilistic model for personalizing the robot's actions based on an individual participant's responses to the robot, and show in simulation that this improves the users' engagement levels compared to models that are not personalized. Together, the results of this work indicate the promise of personalized SAR for helping individuals with cerebral palsy to practice non-verbal communication movements.
\begin{figure*}[t]
\includegraphics[width=\linewidth]{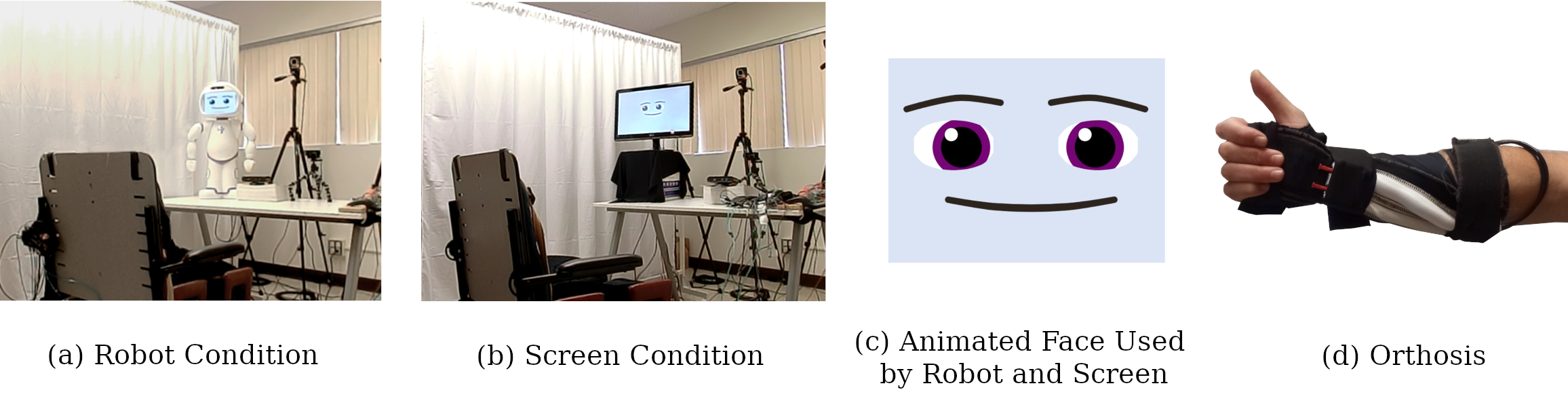}
    % \begin{subfigure}{0.28\linewidth}
    %   \centering
    %     \includegraphics[width=\linewidth]{figures/RobotSetup_corrected.PNG}
    %     \caption{Robot Condition}
    %     \label{fig:setup_robot}
    % \end{subfigure}\hfil % <-- added
    % ~    
    % \begin{subfigure}{0.28\linewidth}
    %     \centering
    %     \includegraphics[width=\linewidth]{figures/ComputerSetup_corrected.PNG}
    %     \caption{Screen Condition}
    %     \label{fig:setup_screen}
    % \end{subfigure}\hfil % <-- added
    % ~
    % \begin{subfigure}{0.2\linewidth}
    %   \centering
    %     \includegraphics[width=\linewidth]{figures/face.png}
    %     \vspace{1cm}
    %     \caption{Animated Face Used By the Robot and Computer}
    %     \label{fig:face}
    % \end{subfigure}\hfil % <-- added
    % ~
    % \begin{subfigure}{0.2\linewidth}
    %   \centering
    %     \includegraphics[width=\linewidth]{figures/orthosis_no_background.png}
    %     \caption{Orthosis}
    %     \label{fig:ortho}
    % \end{subfigure}\hfil % <-- added
    
  \caption{Study setup. The participants used non-verbal gestures to communicate with the robot or computer screen, depending on the experiment condition.  The external 3D camera collected real-time participant hand movement data used by the robot.}
  \label{fig:setup}
\end{figure*}

\section{Related Work}

Engagement is a key factor in measuring the quality of HRI scenarios\cite{oertel2020engagement}, and in particular has been studied extensively as means of maintaining user interest~\cite{celiktutan2018computational}. User-specific perceptual models to identify user engagement have been explored in the context of the autism spectrum disorder \cite{jain2020modeling,rudovic2017measuring,rudovic2019personalized}, where user behavior varies significantly due to personal differences. These personal differences are also present in CP populations, where there is a great variance between individuals in how motor function is impacted. 

Consequently, there has been an emphasis on developing personalized user models to facilitate interactions (see reviews by Clabaugh et al. \cite{clabaugh2019long} and Rossi et al.\cite{rossi2017user}). Personalized interactions have shown promising results in various domains, ranging from rehabilitation \cite{tapus2008user} to robot tutoring systems \cite{leyzberg2014personalizing}, by implementing robot action selection based on personalized user models; however, few have studied engagement dynamics.

A review of several studies \cite{malik2016emergence} concludes that SARs have been effective for clinical populations diagnosed with CP, demonstrating that physical robots can elicit positive responses from users with CP who are performing repetitive physical exercise tasks. Robots as partners in game-like therapeutic physical activities have been shown to create engaging experiences for users and lead to increased motivation \cite{brisben2005cosmobot}. The importance of engagement is emphasized in studies involving movement exercises for CP, and quantitative measures of engagement are well-established for this context \cite{malik2014human}. Given the success of using SARs with this population, we aim to understand how SARs can shape user engagement in practicing repetitive exercises.

\section{Methods}
\begin{figure}[t]
\centering
  \includegraphics[width=0.98\linewidth]{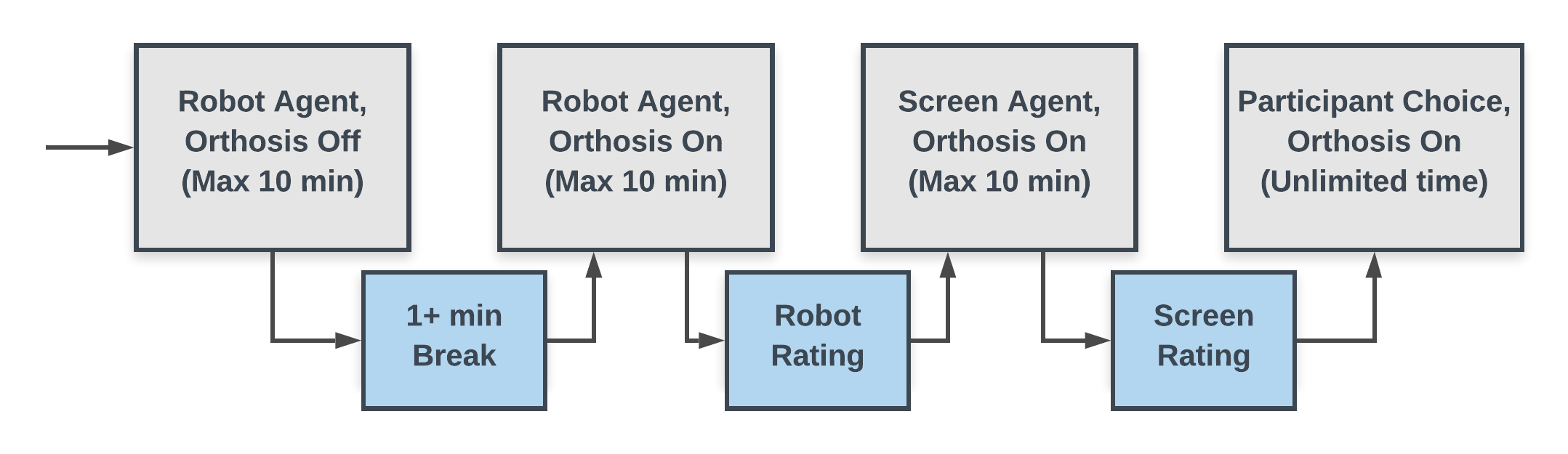}
  \caption{Stages of the within-subject experiment design.}
  \label{fig:studyDiagram}
\end{figure}

\subsection{Study Setup}

The study setup consisted of the participant sitting at a table and facing a computer screen or a robot, both at eye-level, as shown in Figure \ref{fig:setup}.  The experimenter was present in the room for safety and to collect verbal questionnaire data.  Finally, the participant's parent was located in the hallway outside, seated separately and not interacting with the study.

We used the tabletop LuxAI QT robot \cite{luxai}, shown in Figure 1-a; the robot is 25 inches tall, has arms with three DOF arms and a head with two DOF with a screen face. The robot was modified to work with the CoRDial dialogue manager \cite{short2017sprite} that synchronizes facial expressions with text-to-speech. 

The screen condition used a standard 19" monitor that displayed the same simple animated face as the robot's at the same size, as shown in Figure 1-c, and used the same dialogue manager. The robot and screen agent used the same voice, the same facial features, and the same facial expressions, as well as the same machine vision algorithm. The strongly-embodied robot moved and gestured in the shared desk space with the participant, while the weakly-embodied screen was stationary, as shown in Figure 1-b.

The participants wore an orthosis shown in Figure 1-d \\that used fabric-based helical actuators to support the participant's thumbs-up and thumbs-down gestures. The orthosis was controlled by a Beaglebone microprocessor \cite{long_kridner_2019}, actuated with a compressed air tank, and was attached to the participant with Velcro strips for facile donning and doffing \cite{realmuto2019robotic}. The orthosis was worn throughout the session and was not a manipulated variable in the experiment.

The participant's thumb angle was measured by an RGBD camera and transmitted through a ROS network \cite{ros}. A webcam placed on the table in front of the participant captured and recorded the participant's facial expressions. An emergency stop button was provided to the participant for terminating the interaction at any point. 

\subsection{Interaction Design}

At the start of the session, each participant demonstrated a thumbs-up and thumbs-down gesture to generate a baseline for their individual range of motion. Next, the robot explained the number-guessing game, telling the participant to think of a number between 1 and 50, and to communicate the number secretly to the experimenter, by whispering or typing the number on an iPad. At each turn of the game, the robot guessed the number and asked the participant if the guess was correct. The participant answered yes or no by making a thumbs-up or thumbs-down gesture, respectively, using the arm with the orthosis. If the robot guessed incorrectly, it asked if the number was higher than the guess. The participant then answered thumbs-up if the number was higher, and a thumbs-down if the number was lower. The robot ensured that the number of thumbs-up and thumbs-down gestures were approximately equal by tracking the number of each and guessing higher or lower than the target number to keep the counts balanced. The robot continued to guess numbers randomly from a range of decreasing size as it narrowed in on the correct answer. Once the robot correctly guessed the number and the participant signalled with a thumbs-up, the robot asked to play the game again. The participant responded with another thumbs-up or thumbs-down gesture.  

Every time the participant used a thumbs-up or thumbs-down gesture to respond to the robot, the robot responded with feedback that combined verbal, physical, and facial action, based on the quality of the gesture and the history of the participant's gestures.  Specifically, the feedback was a \textit{clarifying} utterance, an \textit{encouraging} utterance, or a \textit{rewarding} utterance, accompanied with a corresponding physical gesture and facial expression. \textit{Clarifying} actions were given when the participant's response was not legible. \textit{Encouraging} actions were given when the angle the participant's thumb made was near their personal baseline value. \textit{Rewarding} actions were given when the participant's thumb angle exceeded their personal baseline value. All verbal, physical, and facial components of these feedback actions were selected randomly from a set of appropriate components for each action, to avoid repetition.

\subsection{Study Design}

% \snnote{it would be good to have a figure that describes the study process}
The study used a within-subjects design shown in the block diagram in Figure~\ref{fig:studyDiagram}; the participants interacted with the robot in a single session that lasted approximately one hour from the participants entering the room to their departure. The session was divided into four blocks, with periods of rest in between. The first three blocks lasted up to 10 minutes each and had the participant play as many games with the robot/screen as desired, while the final block was open-ended, with no fixed duration. Between blocks, the participant rested for at least one minute or until they were ready to move to the next block to mitigate effects of muscle fatigue. The first block served as a practice block to familiarize the participant with the interaction. In that block, the participant interacted with the robot while the orthosis was not powered and thus not assisting their movement. In the second block, the participant interacted with the robot with the orthosis powered on. After the second block, the experimenter verbally administered a questionnaire about the participant's experience with the robot. In the third block, the participant interacted with a computer screen with the orthosis powered on. After the third block, the experimenter verbally administered a questionnaire on the participant's experience with the screen-based agent. The fourth block was optional, and the participant was given a choice of playing with the robot, playing with the screen, or ending the session.

\subsection{Hypotheses}

Since strongly-embodied physical agents have been shown to increase engagement and positive outcomes in therapeutic tasks~\cite{deng2019embodiment,malik2016emergence}, the following hypotheses were tested:

\textbf{H1:} Users with CP will \textbf{prefer} the robot over the screen.

\textbf{H2:} Users with CP will be \textbf{more engaged} when interacting with the robot than the  screen.

% \subsection{Robot and Hardware}
% We used a QT Robot, a humanoid robot developed by LuxAI \cite{luxai}, shown in Figure~\ref{fig:setup}. QTRobot is 25 inches tall, has two 3 degree-of-freedom arms, and has one 2 degree-of-freedom head with a screen-based face. The robot was modified to work with the CoRDial dialogue manager \cite{short2017sprite}, which synchronizes facial expressions with text-to-speech capabilities. In the screen condition, we used a standard monitor that displayed a simple face of the same size as the robot's and used the same dialogue manager. 

% The participants wore an orthosis that used fabric-based helical actuators to support the participant's gestures. The orthosis was controlled by a Beaglebone microprocessor, actuated with a compressed air tank, and was attached to the participant with velcro strips for facile donning and doffing \cite{realmuto2019robotic}. The orthosis was a constant part of all conditions and was not the manipulated variable for this experiment.

% The participant's thumb angle was measured by an RGBD camera and transmitted through a ROS network. The orthosis also connected to the ROS network. A webcam placed on the table in front of the participant captured and recorded the participant's facial expressions. An emergency stop button was provided to the participant for terminating the interaction at any time. 

\begin{figure*}[ht]
\centering
  \includegraphics[width=0.95\linewidth]{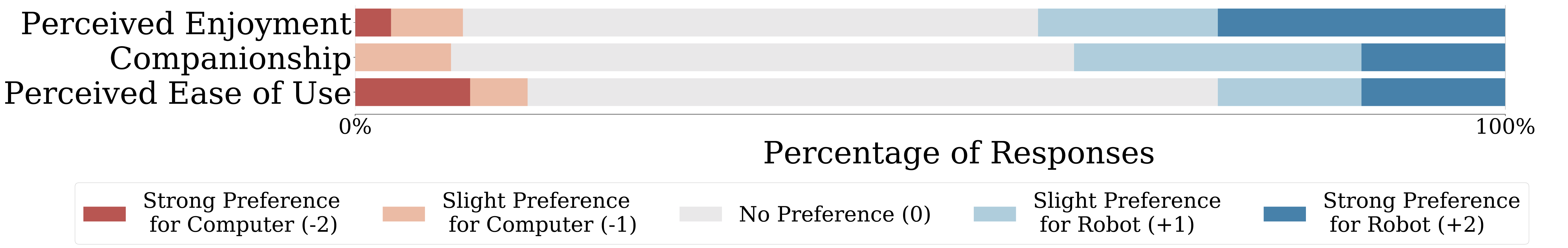}
  \caption{Participant responses to Likert-scale questions, grouped by measured construct.}
  \label{fig:LikertResponses}
\end{figure*}

\subsection{Study Population}
We recruited 10 participants (3 female, 7 male) diagnosed with CP and having symptoms of dystonia in at least one upper limb. The age range of the participants was 9-22 years, with a median age of 15 years. The gender imbalance is representative of the higher prevalence of males in CP populations \cite{johnston2007sex}, and the large age range reflects the challenges of recruitment of this population.  Half of the participants wore the orthosis on their left hand, and the other half wore the orthosis on their right hand. All participants successfully completed the study and were provided with compensation for their time. This study was approved by the University of Southern California Institutional Review Board under protocol \#UP-19-00185.

%The study recruited 10 participants (3 female, 7 male) who had been diagnosed with CP and have symptoms of dystonia in at least one upper limb. The age range of the participants was 9-22 years, with a median age of 15 years. Half of the participants wore the orthosis on their left hand, and the other half wore the orthosis on their right hand. Despite relatively few participants, the population was diverse and representative of the overall population of children with CP. This study and the collection of audio and visual data was approved by the University of Southern California Institutional Review Board under the protocol in \#UP-19-00185.

%--------  Survey Questions Table --------------
\begin{table}[t]
\caption{\label{t:QuestionsAndFactors} Survey questions and associated factors of\\ Companionship (C), Perceived Enjoyment (PE), and \\ Perceived Ease of Use (PEU).}
\begin{adjustwidth}{}{}
    \scriptsize
    \centering
\begin{tabular}{|c|c|}\hline
\textbf{Survey Question} & \textbf{Factor}  \\\hline\hline
\begin{tabular}[t]{@{}l@{}}
How much do you like playing with the \{robot, screen\}?\\
How much do you want to play again with the \{robot, screen\}? \\
How friendly is the \{robot, screen\}?\\
Is the \{robot, screen\} exciting? \\
Is the \{robot, screen\} fun? \\
Does the \{robot, screen\} keep you happy during the game?\\
Is the \{robot, screen\} boring? (inverted)\\
Is playing with the \{robot, screen\} easy?\\
Is communicating with the\{robot, screen\} easy?\\
Is the \{robot, screen\} useful when playing the game?\\
Is the \{robot, screen\} helpful when playing the game?\\
Is playing with the \{robot, screen\} hard? (inverted)

\end{tabular} &
\begin{tabular}[t]{@{}l@{}}
C \cite{lee2005can}\\
C \cite{lee2005can}\\
C \cite{lee2005can}\\
PE \cite{moon2001extending}\\
PE \cite{moon2001extending}\\
PE \cite{moon2001extending}\\
PE \cite{heerink2009}\\
PEU \cite{venkatesh2000determinants}\\
PEU \cite{venkatesh2000determinants}\\
PEU \cite{venkatesh2000determinants}\\
PEU \cite{venkatesh2000determinants}\\
PEU \cite{venkatesh2000determinants}\\
\end{tabular} \\
\hline
\end{tabular}
    \end{adjustwidth}
\end{table}
% --------  End Survey Questions Table --------------

\subsection{Measures}
The participant preference of the embodiment (robot vs. screen) was measured using a three-factor five-point Likert scale, with questions from scales validated in previous works \cite{heerink2009,lee2005can,moon2001extending,venkatesh2000determinants}. The three factors were: perceived enjoyment, companionship, and perceived ease of use. 

The participants' engagement was quantified using identical criteria as in Clabaugh et al. \cite{clabaugh2019long}, which also measures engagement in a game-based interaction. The participant was labelled as engaged if they responded to the robot's question, thought about the correct answer to the question, had a positive facial expression, and was looking at the robot as seen in the auditory and visual data captured by the camera. We represented the level of engagement as a binary variable (engaged/not engaged), as measured by a trained annotator. To ensure consistency, a secondary annotator independently annotated 10\% of the videos selected at random. We measured inter-rater reliability with Cohen's Kappa, and achieved substantial agreement of $k=.73$, corresponding to an agreement on 86\% of videos, similar to other works in engagement \cite{rudovic2018personalized,clabaugh2019long}.

\section{Study Results}

\subsection{User Preference}
Embodiment preference was determined by the difference in ratings between corresponding questions for the robot and screen conditions. The specific questions are shown in Table~\ref{t:QuestionsAndFactors}. The combined responses for all factors are shown in Figure~\ref{fig:LikertResponses}. We found high internal consistency for all factors: Perceived Enjoyment ($\alpha=.94$), Companionship ($\alpha=.91$), and Ease of Use ($\alpha=.89$). We evaluated significance with a Wilcoxon Signed-Rank Test and found a significant preference for the robot over the screen in factors measuring Companionship ($Z=9.0$, $p = .026$) and Perceived Enjoyment ($Z=23.5$, $p = .018$). We found no significant differences in Ease of Use ($Z=52.0$, $p=.399$) and attribute this to the fact that both embodiments used the same vision system, which suffered from perceptual errors (such as failing to detect the participant's off-camera thumb angle) about 20\% of the time. We additionally note that many of the responses showed no preference for either the robot or screen due to the tendency of the participants to respond with similar values for all questions. The results therefore partially support H1, indicating that participants somewhat preferred to interact with the robot over the screen.
 
\subsection{Choice Condition}
In the final study block, participants could choose to play a game with the robot, play another game with the screen, or stop playing and end the session. One participant chose to play with the robot, five participants chose to play with the screen, and four participants chose to stop playing. While the sample size is too small to draw any conclusions, the confounding factors include the possibility that participants were too fatigued to continue playing or reluctant to require work of the experimenter to exchange the embodiments. Several participants expressed these sentiments while doffing the orthosis at the conclusion of the experiment.

\subsection{Engagement}
 We found no significant differences in the participants' engagement between the two conditions, which does not support H2; our post-hoc analysis shows that there were individualized differences in how engagement changed in response to the robot's actions. We discuss those differences next in the context of personalizing the interaction.
 
\section{Modeling Engagement}
\label{sec:modeling}
We first explored whether there were differences across users in how engagement changed in response to the robot's actions. We modeled the evolution of engagement as a Markov chain, where engagement is a binary state variable that changes stochastically in discrete time-steps, after each of the robot's actions. 

We define a transition matrix $T$ that specifies how engagement $s \in S$ changes over time $T: S \times A \rightarrow \Pi(S)$. Since the change depends on the robot's action (\textit{clarify, encourage} or \textit{reward}), we parameterized the transition matrix by the robot's action $a \in A$. 

\subsection{Learning Personalized Models}
\label{subsec:learning}
To learn personalized engagement models, the first step is to represent how likely a participant is to become engaged or disengaged given a robot's action. We captured this with the transition matrix of the Markov chain. For each participant in the user study, we computed a transition matrix using maximum likelihood estimation from the sequence of the annotated engagement values. 

We next explored whether participants cluster in terms of similar reactions to the robot's actions. Previous work~\cite{nikolaidis2015efficient} has shown that users can be grouped based on their preference on how to perform a collaborative task with a robot. We used a similar approach in the context of social interaction: we clustered participants from the study based how their engagement changed in response to the robot's actions.

We converted the transition matrices to vectors, then computed the distance between vectors using cosine similarity. We then performed hierarchical clustering \cite{mullner2011modern}, by iteratively merging the two most similar vectors into a cluster. The merged vector was formed by averaging the values of the two vectors. We selected the final number of clusters, so that each cluster contained at least two individuals. We transformed the vector of each cluster back to a transition matrix that specified how engagement changed for participants of that cluster.

We clustered participants at two different resolutions: 1) based on the user as a whole and 2) based on the users' response to each of the robot's three possible actions.  The first clustering, based on each participant's holistic response to to robot actions, resulted in matching each participant to one cluster. We call this \textit{participant-level clustering}. The second clustering, based on each participant's responses to each of the robot's action separately, required three different clustering iterations, one for each action, and resulted in having each participant matched to three clusters, one for each action. We call this \textit{action-level clustering}.

\begin{figure}[t]
\centering
  \includegraphics[width=0.95\linewidth]{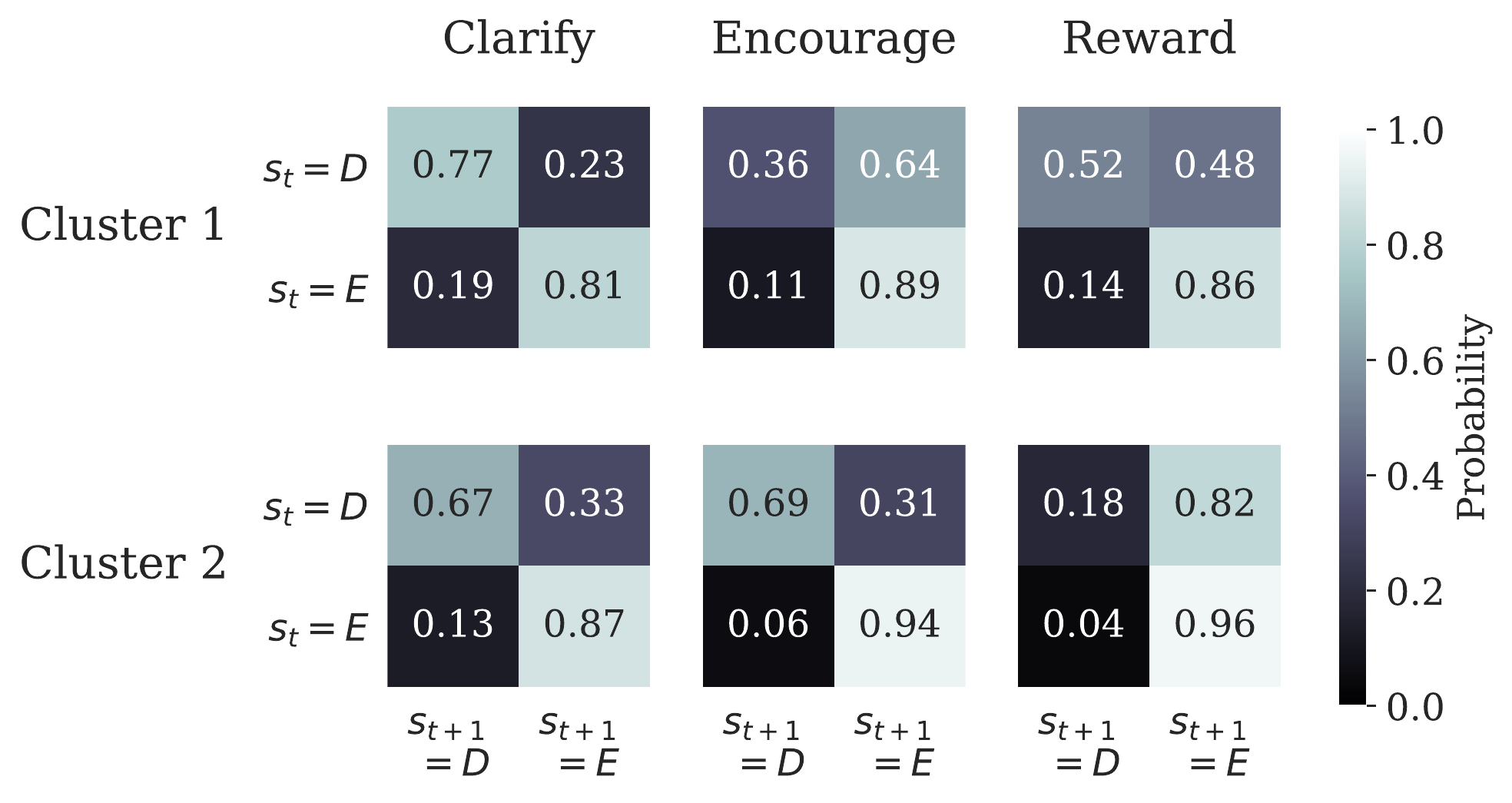}
  \caption{Transition matrices of the two clusters found in the participant-based clustering method. Each matrix specifies the probability of becoming engaged (E) or disengaged (D) at the next time-step, given the current state.}
%   \snnote{This needs a description and a color bar.}
  \label{fig:participantCluster}
  
\end{figure}

\subsection{Cluster Identification} \label{clusters}

%At the participant level, a user was defined as a vector of all of their transition probabilities. These vectors were then clustered hierarchically by cosine similarity by iteratively merging the two vectors that are closest together until there is only one vector left. At each iteration, the merging of two vectors results in a parent vector that is equidistant to the two child vectors.
% \snnote{this needs more information}. 
%This builds a tree structure where $n-1$ iterations from forming the root results in $n$ clusters. The number of clusters were selected such that each cluster contained at least two children \snnote{not sure what this means}. 

Using the participant-level method, we found two main clusters, shown in Figure~\ref{fig:participantCluster}. In the first cluster, the \textit{encourage} action had a greater likelihood of causing the participant's next state to be Engaged (E) than the \textit{reward} action. In the second cluster we observed the opposite effect: the probability of changing from Disengaged (D) to Engaged (E) is lower for the \textit{encourage} action than for the \textit{reward} action. We observe that the \textit{clarify} action has a small effect on changing a participant's engagement state in both clusters. Seven participants belonged to the first cluster, and three participants belonged to the second cluster. There were no clear factors that lead to the makeup of the participants in the clusters based on the background information collected in the study.

\begin{figure}[t]
\centering
  \includegraphics[width=0.95\linewidth]{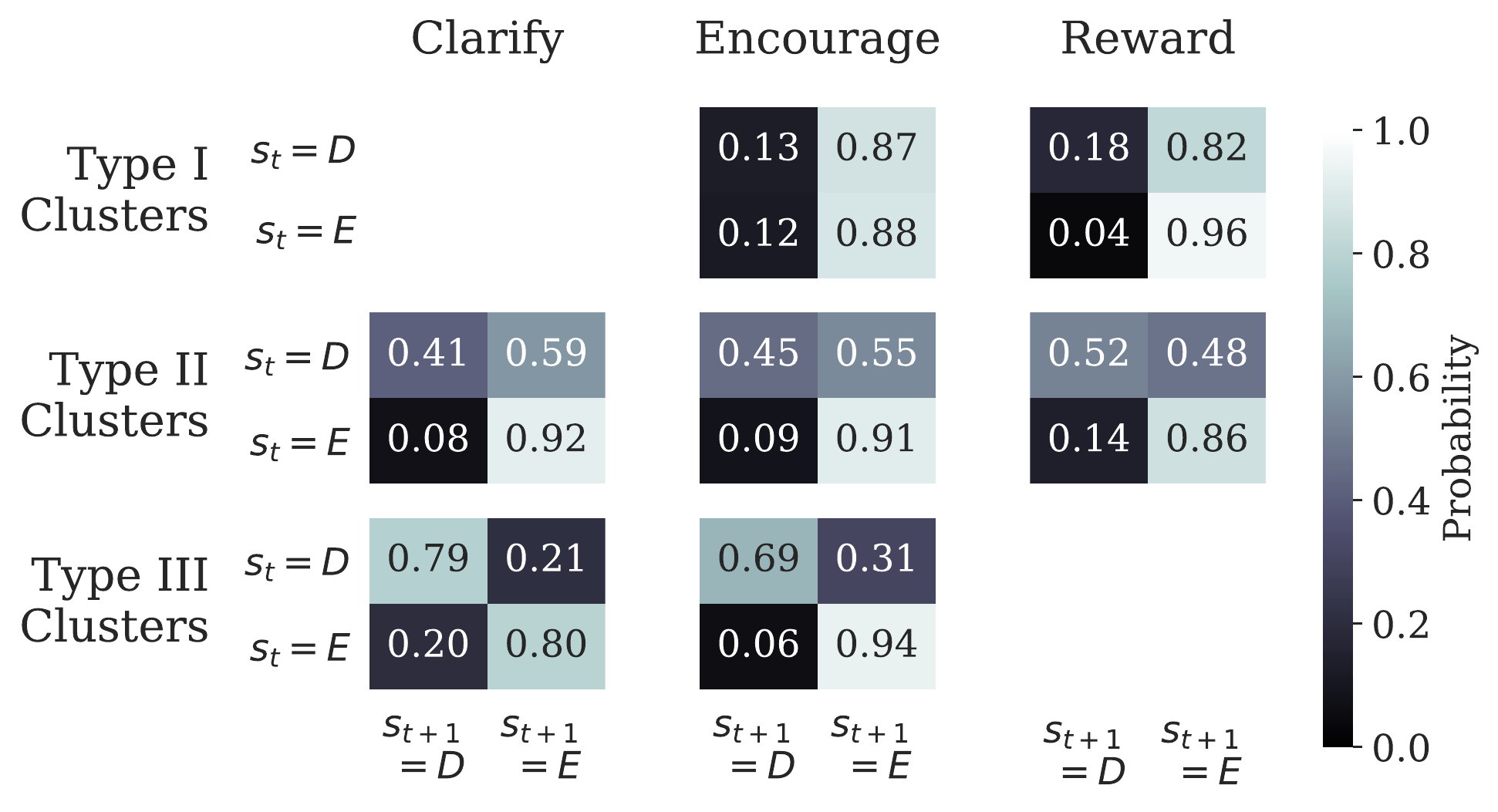}
  \caption{Transition matrices of different clusters found in the action-based clustering method. Each matrix specifies the probability of becoming engaged (E) or disengaged (D) at the next time-step, given the current state.} 
%   \snnote{the font in figures should be of the same size as th caption}
\vspace{-.4cm}
  \label{fig:actionCluster}
\end{figure}

The action-level clustering method generated separate clusters for each action independently (Figure~\ref{fig:actionCluster}). Thus, a single participant is described as being a part of three action-level clusters. We observed three different types of matrices across the different actions: 
% Type I indicates a high probability of the participant becoming engaged regardless of their previous state, Type II has an approximately equal probability of becoming engaged or remaining disengaged if the participant was previously disengaged, and Type III features participants who are most likely to remain in the same state. 

\begin{itemize}
    \item Type I indicates a high probability of the participant becoming Engaged, regardless of their previous state. 
    \item Type II has an approximately equal probability of becoming Engaged or remaining Disengaged, if the participant was previously Disengaged. 
    \item Type III features participants who are most likely to remain in the same state. 
\end{itemize}

We found that the {\it clarify} action generated only Type II and III clusters, since most participants' engagement did not change based on that action, with three participants belonging to the Type II cluster and seven participants belonging to the Type III cluster when conditioning on the {\it clarify} action. The {\it reward} action generated Type I and Type II clusters, since most participants became Engaged after a \textit{reward} action. Three participants belonged to the Type I cluster and seven participants belonged to Type II cluster. This finding supports previous work~\cite{fogg1997silicon} that showed positive reinforcement improving participants' engagement in computer-based animal guessing games. 

Only the {\it encourage} action generated clusters of all three types. The {\it encourage} action had three participants that responded in alignment with the Type I cluster, five participants that aligned with the Type II cluster, and two participants that aligned with the Type III cluster. 

We investigated whether the composition of the participants in each cluster was related to the demographic information we collected. Specifically, we analyzed cluster composition with a multinomial logistic regression of age, gender, and handedness onto cluster type and found no significant differences in composition between any clusters at either participant or action levels. Qualitatively, age was a minor component in the Encourage clusters; older ages appeared to be more associated with the Type II cluster (median age 16), whereas younger participants either fell into Type I (median age 14) or Type III (median age 10.5) clusters.

% \begin{figure}[t]
% \centering
%   \includegraphics[width=0.65\textwidth]{figures/UserInference.png}
%   \caption{Histogram of Change in Time Engaged over Random Action Selection}
%   \label{fig:userEngagement}
% \end{figure}

\begin{figure*}[t]
\centering
\begin{tabular}{cc}
\hspace{-1.5em}

\begin{subfigure}[b]{.33\linewidth}
\centering
  \begin{tabular}{cc}
  \resizebox{1.0\linewidth}{!}{
  \includegraphics[width=\linewidth]{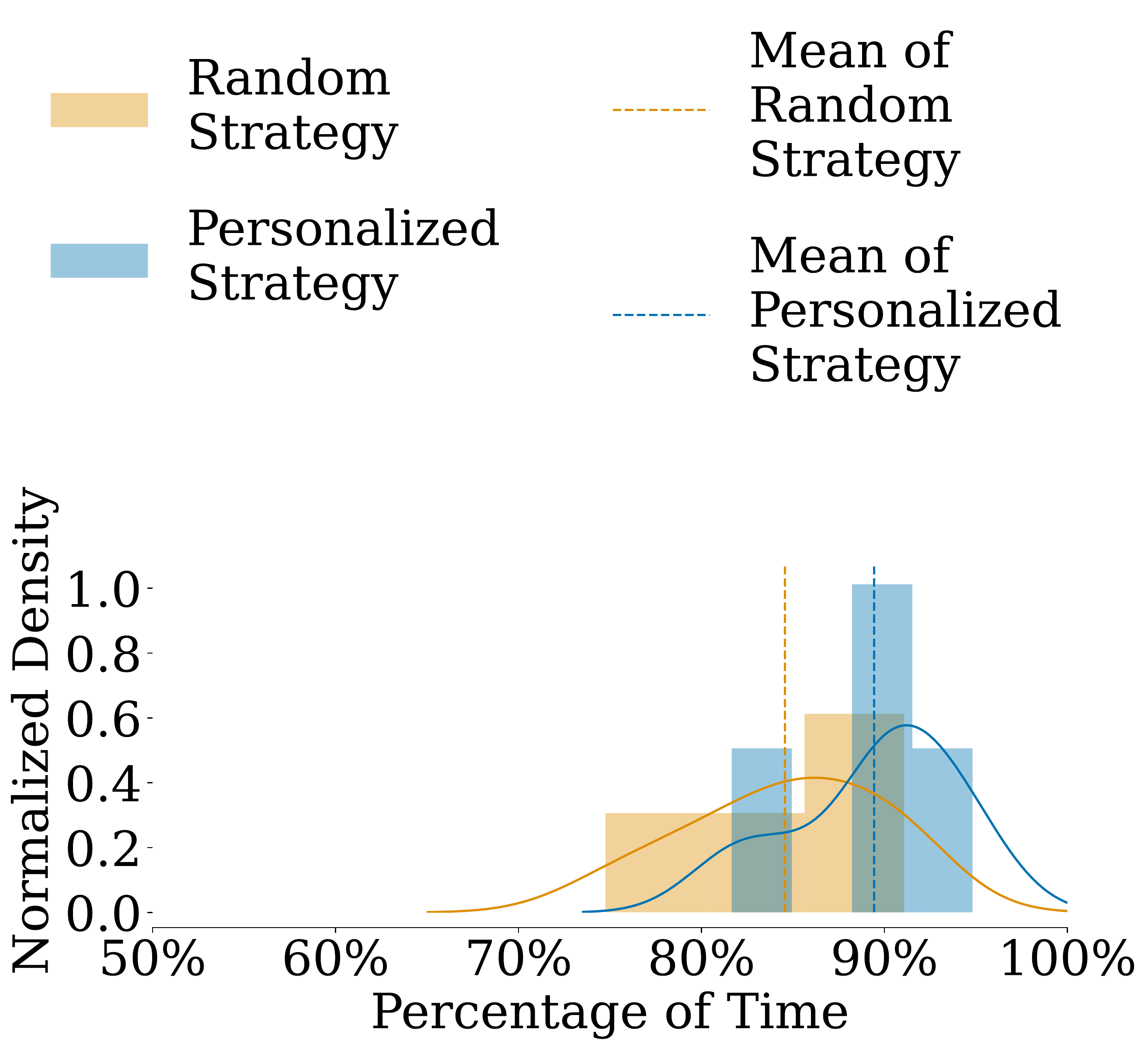}
   }
   \end{tabular}
 \label{fig:plotT1}
 \caption{Correct User Inference vs. Random Selection}
 \end{subfigure}
 
 \begin{subfigure}[b]{.33\linewidth}
\centering
  \begin{tabular}{cc}
 \resizebox{1.0\linewidth}{!}{
  \includegraphics[width=\linewidth]{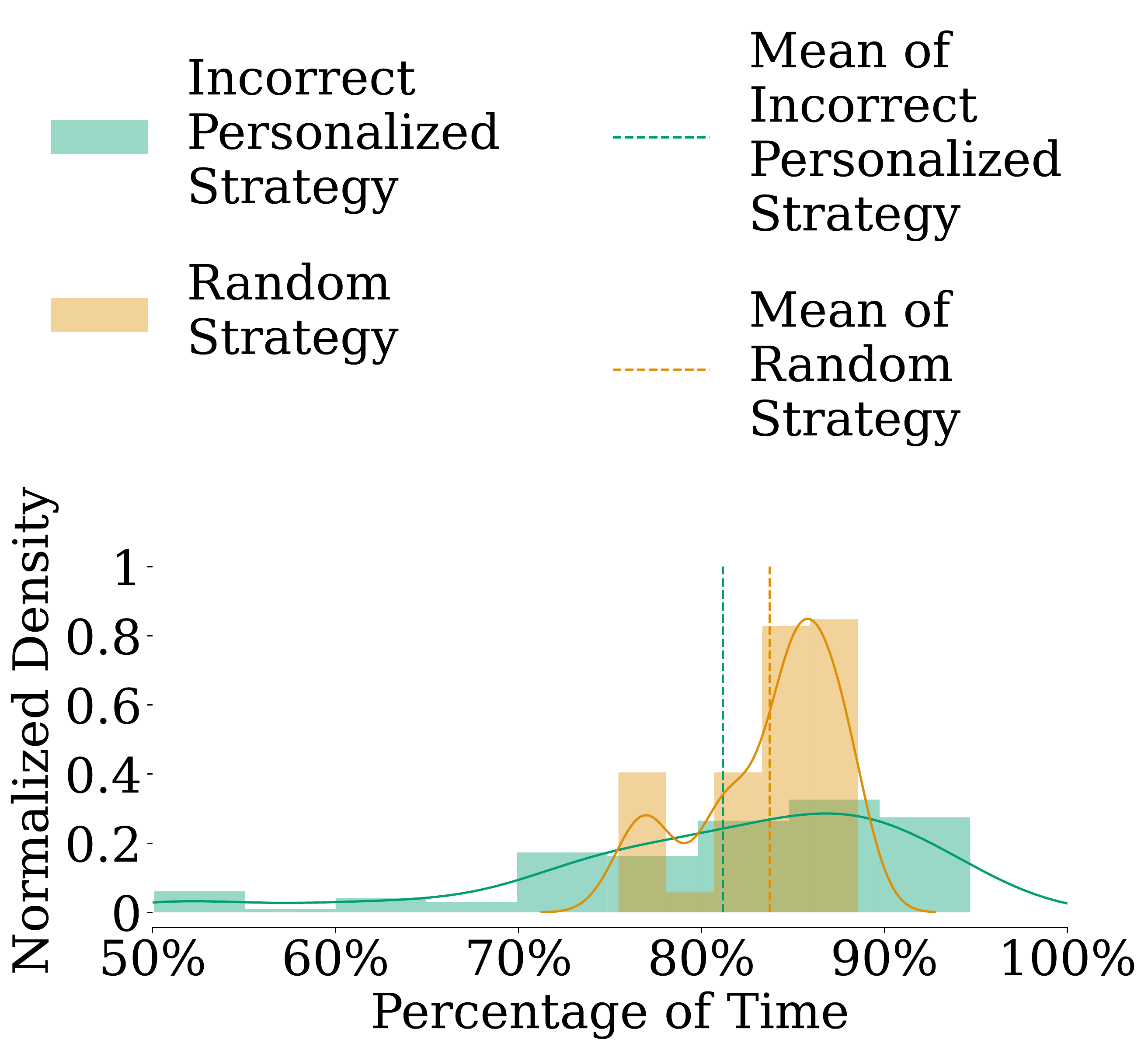}
   }
   \end{tabular}
 \label{fig:plotT0}
 \caption{Incorrect User Inference vs. Random Selection}
\end{subfigure}
%&
\begin{subfigure}[b]{.33\linewidth}
\centering
  \begin{tabular}{cc}
  \resizebox{1.0\linewidth}{!}{
  \includegraphics[width=\linewidth]{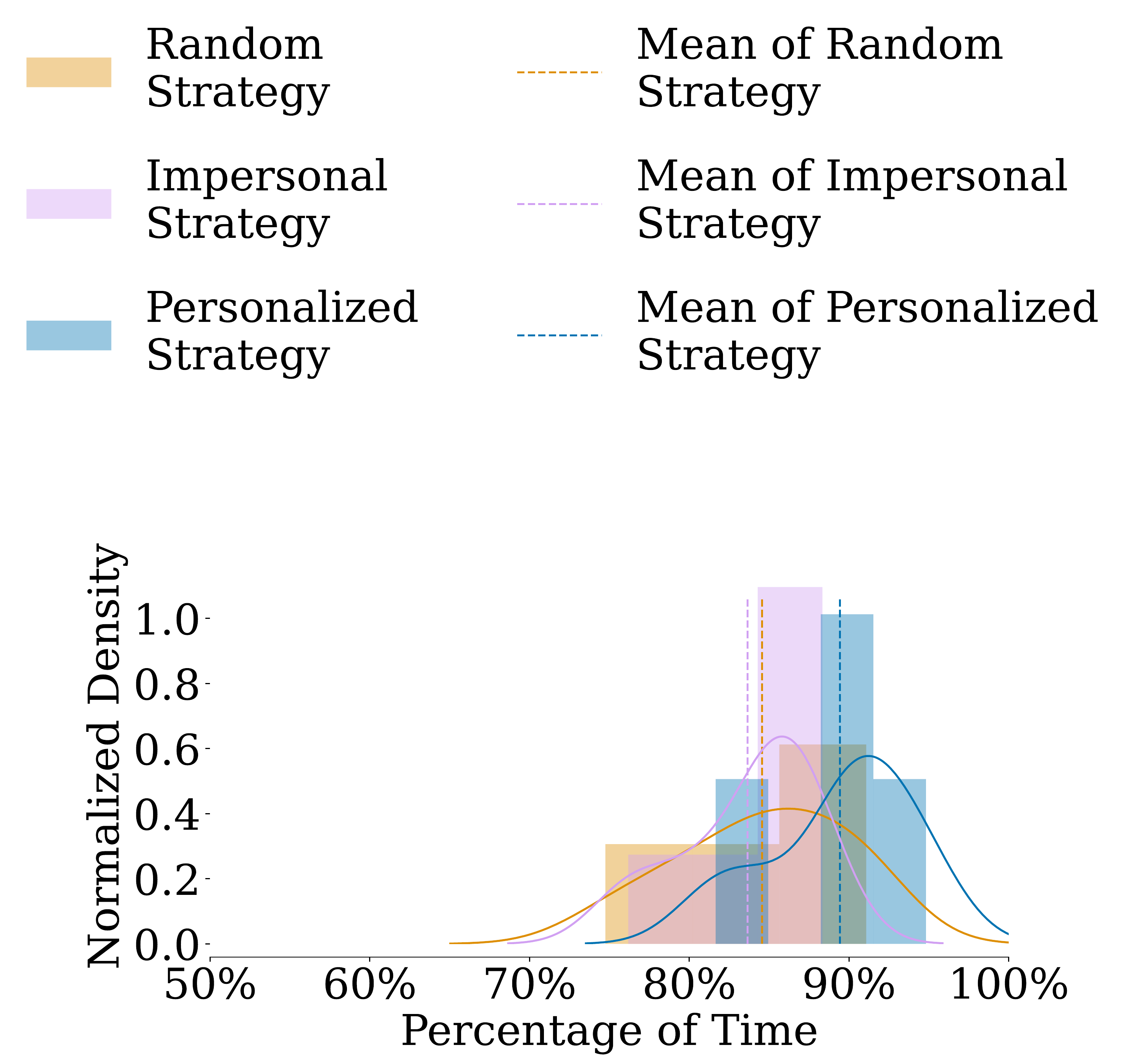}
   }
   \end{tabular}
 \label{fig:plotT2}
 \caption{Comparison of all strategies}
\end{subfigure}
%&

%&
\end{tabular}
\caption{
Percentage of time that modeled users were engaged for different methods of robot action selection. Selecting actions based on the correct user clusters (a) keeps users more engaged, however selecting actions on incorrect user models (b) has an adverse effect. Considering the users as one group (c) performs similarly to the random baseline.
% We compare maximizing engagement based on action-level clusters or (a) random action selection, and (b) maximizing engagement over a single cluster. In (c) we show the effect of maximizing engagement based on a cluster different than the correct cluster of the modeled user. 
}
\label{fig:simulatedResults}
\end{figure*}

\subsection{Personalizing Robot Actions}
\textit{If a robot knows the participant's cluster and adapts its actions to maximize engagement, to what extent does this improve the participant's engagement?} Following prior work in simulating users based on personas \cite{andriella2019}, we show the benefit of using a personalized engagement model by modeling users based on the data from the user study. We focus on the action-level clustering method, since it generates different clusters for each robot action, resulting in a higher resolution model than the participant-level clusters using the same amount of data.  

% We modeled users based on each participant's transition matrices described in Section~\ref{subsec:learning}. Given a modeled user's engagement level and a robot action, we generated the next user's state by sampling from the transition matrix. This new state is then used as the ground truth user state to choose the next action 

We modeled users based on each participant's transition matrices described in Section~\ref{subsec:learning}. At each timestep, we have the user's current engagement state and the set of the ground-truth transition matrices for each action from the study. When the robot takes an action, the user's next engagement state is sampled from the probability distribution of the corresponding action and the current engagement state. This process is repeated for 100 timesteps, as determined by the number of turns in the in-person study. We additionally average our results over 100 runs for each participant to mitigate the random effects of the simulation and converge to the true mean of the engagement level over the course of the simulation. 

Our strategy for selecting actions in the simulation is to maximize the likelihood of the user becoming engaged based on the estimated user clusters. For instance, if a user is said to be Type II for {\it clarify}, Type I for {\it encourage} and Type II for {\it reward}, and the participant is currently disengaged (D), then the robot would choose the {\it encourage} action, since the participant would become engaged (E) with probability 0.87 (compared to 0.48 for {\it reward} and 0.59 for {\it clarify}). These estimated clusters, however, are distinct from the true clusters used to simulate the users: for example, a user may truly be associated with the type II cluster for {\it encourage} actions, but we may erroneously select actions as if that user were associated with a type I cluster for the {\it encourage} action. We additionally imposed a 0.2 probability of the robot taking a {\it clarify} action no matter what state the participant was in to account for incorrect or illegible gestures; similar to the rate that was observed in the in-person study. 

% These estimated clusters, however, are distinct from the true clusters used to simulate the users: for example, a user may truly be associated with the type II cluster for {\it encourage} actions, but we may erroneously select actions as if that user were associated with a type I cluster for the {\it encourage} action.

We computed two baselines: 1) the robot selects actions randomly, choosing {\it encourage} or {\it reward} action with equal probability; and 2) the robot selects actions to maximize engagement based on the transition matrix computed from the maximum likelihood estimate of all the participants. The second baseline, which we call the ``impersonal strategy", is equivalent to having one cluster for every action; it does not account for individual differences. 

 Fig.~\ref{fig:simulatedResults} shows the average time the modeled users spent being Engaged in the activity for each condition. A two-tailed paired t-test showed that modeled users spent significantly more time being Engaged when the robot selected actions given their cluster compared to taking random actions ($t(11)=9.370$, $p<.001$). However, if the robot had a flawed model of the user, the modeled users were significantly less engaged over the course of the interaction compared to randomly selecting actions ($t(131)=-4.692$, $p<.001$). This result highlights the trade-offs that personalization may bring, especially in low-data scenarios.

The second baseline treated users as coming from one cluster. The personalized strategy significantly outperformed the impersonal strategy ($t(11)=4.984$, $p<.001$). Furthermore, we cannot say that the impersonal strategy performed any differently than randomly selecting actions ($t(11)=-.656$, $p=.525$). This shows the importance of algorithmic design in interaction, and how incorrect assumptions can lead to data-driven models that are ineffective.

\section{Design Insights}

From the study conducted with a participants with CP, we developed the following design insights that may be useful when designing personalized algorithms for end-users.

{\it The distribution of interaction preferences is often unbalanced.} The clusters formed from this study revealed skews in the number of participants, most commonly with the majority cluster being composed of seven of the ten participants. This highlights the importance of understanding the lower-probability modes in which users interact with an implemented system through the use of qualitative tools such as user personas \cite{andriella2019}. These are especially helpful when there are no apparent differences between the clusters.
% Importantly, the different clusters of user preferences are often unbalanced, highlighting the importance of qualitative interpretations of different preferences through design tools such as user personas.

{\it Clustering user responses helps to reduce the design space.} Our clusters reveal three main types of responses to each action. Users found each action as either highly engaging (Type I), sometimes engaging (Type II), or having no effect on engagement (Type III). Interestingly, we did not see actions that were highly disengaging or caused the participant to switch states with high probability. This indicates that those types of interaction are less common, and therefore less focus can be placed on considering how those cases would affect the interaction.
% We show that adopting a robot-centric approach of clustering on robot actions, as opposed to a user-centric approach of clustering user responses as a whole is able to better personalize to individuals, even with limited data.

{\it Certainty in user cluster or persona is critical in personalized algorithms.} Our results show that misclassification of a user's cluster or persona leads to lower engagement than random selection. To build effective systems, it is critical to be certain that the inferred user's classification is correct. A system that personalizes to a given user should consider the risk of misclassification in selecting the level of certainty that is required to make decisions in an interactive scenario.

\section{Limitations and Conclusion}

Our results are limited by our small sample size, resulting from the practical challenges of recruiting participants with CP. Applying our clustering approach with more participants will likely result in more nuanced representations of user engagement levels. Our method also carries the inherent limitations of Markov-based models: it does not account for effects of the history of interaction, such as fatigue, or aspects of the interaction that are not modeled, such as participants getting distracted by other events in their environment. 

Additionally, our user models are based on the assumption that users can be associated with a previously known classification; they do not account for new, previously unseen classifications. Implementation in a real-world setting would also require correct inference of the user's engagement level in real time, as well as accurate identification of the user's classification. In fact, our user models show that incorrect inference results in worse performance than random robot action selection.

This work introduces socially assistive robotics to the context of communicative gesture practice for users with cerebral palsy. Our user study shows that participants with CP preferred to interact with a socially assistive robot compared to a screen-based agent. While we did not observe significant differences in user engagement overall, our post-hoc analysis showed that there are nuanced differences between modes of participants: some participants became more engaged after the robot gave encouraging feedback, while others responded better to rewarding feedback. We show that understanding how a user will react to different robot actions can be leveraged to design a more engaging experience.

% Our models of users show the promise of personalizing the robot's actions based on classification. Our study additionally suggests important design considerations for personalizing interactions for different users. This work introduces socially assistive robotics to the context of communicative gesture practice for users with cerebral palsy and brings new insights about how user engagement may improve from different types of feedback. and brings new insights about how user engagement may improve from different types of feedback

\bibliography{Submission}

\end{document}